\documentclass{article}

\usepackage[nonatbib, final]{neurips_2021}

\usepackage[utf8]{inputenc} 
\usepackage[T1]{fontenc}    
\usepackage{hyperref}       
\usepackage{url}            
\usepackage{booktabs}       
\usepackage{amsfonts}       
\usepackage{amsmath}        
\usepackage{nicefrac}       
\usepackage{microtype}      
\usepackage{xcolor}         
\usepackage{mathtools}

\usepackage{caption}
\usepackage{subcaption}

\usepackage{tikz}
\usetikzlibrary{positioning, arrows}

\usepackage[ruled,vlined]{algorithm2e}
\usepackage{bm}

\usepackage{siunitx}

\usepackage[
backend=bibtex,
maxbibnames=20,
sorting=none,
date=year,
doi=false,
isbn=false,
url=false
]{biblatex}
\newcommand{\E}{\ensuremath{\mathbb{E}}}				
\newcommand{\x}{\ensuremath{\bm{x}}}					
\newcommand{\xhat}{\ensuremath{\bm{\hat{x}}}}		    
\newcommand{\X}{\ensuremath{\bm{X}}}					
\newcommand{\z}{\ensuremath{\bm{z}}}					

\newcommand{\enc}{\ensuremath{enc_\phi}}				
\newcommand{\dec}{\ensuremath{dec_\theta}}			
\newcommand{\cox}{\ensuremath{cox_\psi}}	  			

\DeclarePairedDelimiterX{\infdivx}[2]{(}{)}{%
	#1\;\delimsize\|\;#2%
}
\newcommand{\kl}[2]{\ensuremath{\mathbb{KL}\infdivx{#1}{#2}}} 
\newcommand{\loss}[1]{\ensuremath{\mathcal{L}_{#1}}}		

\newcommand{\pinf}{\ensuremath{q_\phi(\bm z \mid \bm x)}}			

\newcommand{\lossbvae}{\ensuremath{\loss{\beta {\scriptscriptstyle - VAE}}}}	

\newcommand{\gauss}{\ensuremath{\mathcal{N}}}				

\newcommand{\eqendc}{\,\text{,}} 

\newcommand{\grahl}{HazardWalk}

\title{Towards modelling hazard factors in unstructured data spaces using gradient-based latent interpolation}

\author{%
  Tobias Weber\\
  Department of Statistics\\
  LMU Munich\\
  \texttt{tobias.weber@stat.uni-muenchen.de} \\
   \And
  Michael Ingrisch\\
  Department of  Radiology\\
  LMU Munich\\
  \texttt{michael.ingrisch@med.uni-muenchen.de} \\
  \And
  Bernd Bischl\\
  Department of Statistics\\
  LMU Munich\\
  \texttt{bernd.bischl@stat.uni-muenchen.de} \\
   \And
  David R\"ugamer\\
  Department of Statistics\\
  LMU Munich\\
  \texttt{david.ruegamer@stat.uni-muenchen.de} \\
}

\begin{document}

\maketitle



\begin{abstract}
    The application of deep learning in survival analysis (SA) allows utilizing unstructured and high-dimensional data types uncommon in traditional survival methods. This allows to advance methods in fields such as digital health, predictive maintenance, and churn analysis, but often yields less interpretable and intuitively understandable models due to the black-box character of deep learning-based approaches.  
    We close this gap by proposing 1) a multi-task variational autoencoder (VAE) with survival objective, yielding survival-oriented embeddings, and 2) a novel method \grahl{} that allows to model hazard factors in the original data space.  \grahl{} transforms the latent distribution of our autoencoder into areas of maximized/minimized hazard and then uses the decoder to project changes to the original domain. Our procedure is evaluated on a simulated dataset as well as on a dataset of CT imaging data of patients with liver metastases.
\end{abstract}



\section{Introduction}
 
Survival analysis (SA) is indispensable for estimating factors of increased risk, e.g., for an infection of a contagious infection or the onset of a disease. Survival is also an important clinical endpoint for assessing safety and efficacy of cancer therapies \parencite{Delgado2021}. Despite its importance and practical relevance, modelling survival times remains challenging due to the often complex data generating process, involving censoring, truncation, time-varying features, recurrent events or competing risks. While there exist frameworks such as \cite{Bender2020, Kopper2020a} that reduce the complexity of modelling survival related tasks by relating SA to a Poisson regression using a special data transformation \parencite{Friedman1982}, many recent approaches adapt the semi-parametric Cox proportional hazard (PH) model \parencite{Cox1972}. Instead of relating directly to the survival probability, the Cox PH models the hazard rate, which is the instantaneous risk of experiencing an event (disease, death, etc.).
Modern non-linear adaptions of the Cox PH model employ neural networks and advanced optimizers \parencite{Katzman2016, Luck2017}. Other approaches strive to model the hazard at every possible point in time using a discretized output time domain \parencite{Lee, Lee2020, Ren2018}. More recently, neural ordinary differential equations for use in SA were applied successfully \parencite{Groha2020}.
Generative models and survival have been investigated in \cite{Palsson2019}, who use an autoencoder in a semi-supervised setting to encode survival information and classify between long- and short-term survivors. Moreover, \cite{Bello2019} regularize a denoising autoencoder with a survival loss to perform cardiac motion analysis.

SA is predominantly applied for tabular data. A few exceptions exist (see, e.g., \cite{Haarburger2018}), but a majority of techniques is not designed for unstructured or high-dimensional data. Modern deep learning approaches address this problem, but in contrast to existing SA techniques, do not have a straightforward model interpretation -- a crucial aspect in digital health and medical decision making. This interpretation and understanding is equally important for generative models, yet more intricate and especially challenging in the light of SA. 
We close this gap by proposing a two-step procedure combining ideas of deep generative models and counterfactuals \parencite{Verma2020}. In a first step, we estimate an inherently interpretable latent representation of the unstructured data source using a multi-task variational autoencoder (VAE; \cite{Kingma2014}) with regularizing survival objective. In the second step we help the practitioner to understand the modelled relationship of this latent space and the survival outcome using our novel \textit{\grahl} method. While our autoencoder architecture allows modelling and isolating survival-relevant factors, the \grahl{} can be used to traverse in the latent space guided by the survival-objective and utilizes the generative decoder to project back its reasoning to the original data space (e.g., by an respective visualization in the image).


\section{Methods}


\paragraph{Autoencoder with survival downstream objective}

Our proposed architecture consists of a probabilistic encoder \enc\ and a decoder \dec\ with parameters $\phi$ and $\theta$, respectively. For a sample $\x \in \mathcal{X}$ from the sample space $\mathcal{X}$, the approximate inference distribution \pinf\ of the encoder is given by $\gauss(\bm{\mu}_\phi, \text{diag}(\bm{\sigma}^2_\phi))$ where $\bm{\mu}_\phi, \bm{\sigma}_\phi = \enc(\x)$. By sampling $\z \sim q_\phi(\bm z \mid \bm x)$, a reconstructed version \xhat\ of the data sample $\x$ can be produced by \dec. In order to learn this reconstruction, we optimize the evidence lower bound (ELBO)
\begin{equation} \label{eq:theo-vae-beta}
	\lossbvae(\theta, \phi; \x) =  - \E_{\z \sim \pinf} \left[ \log p_\theta(\x \mid \z) \right] + \beta \kl{q_\phi(\bm z \mid \bm x)}{p (\z)} \eqendc
\end{equation}
where $\log p_\theta(\x | \z)$ accounts for the reconstruction and $\beta$ is an additional parameter to control the impact of the KL-divergence $\mathbb{KL}$ between $q_\phi(\bm z | \bm x)$ and the Gaussian prior $p (\z)$ \parencite{Burgess2018}.

While the plain ($\beta$-)VAE works well for reconstruction, it does not learn concepts to emphasize survival-relevant information in \x.
We therefore propose to extend the ($\beta$)-VAE to a multi-task model with survival downstream objective by adding a network head \cox\ with parameters $\psi$ to the latent bottleneck.
This leads to a latent space guided by survival information. 
The task of \cox\ is to predict the log-hazard rate or risk score $r \in \mathbb{R}$ with $r = \cox(\z)$ for $\z \sim q_\phi(\bm z \mid \bm x)$.
In SA, the hazard $h(t \mid \x)$ describes the instantaneous risk of experiencing the event of interest exactly at a specific time point $t$. Following the original Cox-PH model, the hazard can be decomposed as $h(t \mid \x) = h_0(t) \exp(r)$, where $h_0(t)$ is a feature-independent baseline hazard and $\exp(r)$ serves as a feature-dependent scaling factor based on the risk score $r$.
Let \X~be a dataset of $n$ observations $\bm{x}_i$ with corresponding event times $t_i$ for $i=1,\ldots,n$. For a predicted risk score $r_i = \cox(\z_i)$ based on $\z_i \sim \enc(\x_i)$, the corresponding objective function is given by the negative partial Cox log-likelihood (see, e.g., \parencite{Katzman2016}):
\begin{equation} \label{eq:meth-coxvae-coxloss}
	\mathcal{L}_{\scriptscriptstyle Cox}(\phi,\psi; \X) = - \frac{1}{\sum_{i=1}^{n} \delta_i} \sum_{i=1}^{n} \delta_i \left( r_i - \log \sum\limits_{j: t_j \geq t_i} \exp(r_j)  \right) \eqendc
\end{equation}
where $\delta_i = 0$ if the $i$th observation is censored, else $\delta_i = 1$. By regularizing the autoencoder with the survival downstream task, latent features are forced to learn a representation that is inherently interpretable from a survival point of view. The joint objective of our multi-task approach is given by
\begin{equation} \label{eq:meth-total-loss}
	\mathcal{L}(\phi, \theta, \psi; \X) = \frac{\tau}{n} \sum_{i=1}^n \left( \lossbvae(\phi, \theta; \x_i) \right) + (1 - \tau) \mathcal{L}_{\scriptscriptstyle Cox}(\phi,\psi; \X) \eqendc
\end{equation}
with $\tau\in[0,1]$ controlling the influence of each loss term in the convex combination of both. In our experiments, a value of $\tau = 0.5$ turns out to give both good reconstructions and survival-oriented latent spaces.


\paragraph{Hazard guided walk}

In order to visualize the impact of survival-relevant dimensions in the latent space, we propose the gradient-based hazard latent walk (\textit{\grahl}). \grahl{} is a new method to traverse the latent space based on the estimated hazards by \cox\ and utilizes \dec\ as  generative model to project these changes to the original data space, making important factors on the overall survival directly observable. The latent variational distribution of the VAE thereby ensures smooth transitions in $\mathcal{X}$. 
First, the latent feature distribution is obtained by forwarding a sample \x\ through \enc, yielding parameters $(\bm{\mu}_\phi, \bm{\sigma}_\phi) = \xi$. 
The goal is now to obtain the expected gradient of the hazard rate w.r.t. the whole latent distribution, which points in the direction of maximal hazard rate increase.
We approximate this using Monte Carlo sampling with
\begin{equation} \label{eq:meth-grahl-grad}
     \mathbb{E}_{\pinf} \left[\nabla_{\xi} \exp(\cox(\z)) \right] = \int \nabla_{\xi} \exp(r) \pinf \,d\z \approx \frac{1}{B} \sum_{b=1}^B  \nabla_{\xi} \exp(r_b)
\end{equation}
based on samples $\bm{z}_b \sim q_\phi$. By updating the parameters of \pinf\ in the direction of \eqref{eq:meth-grahl-grad} and repeating this procedure for a pre-specified amount of iterations, we shift and scale the latent distribution into areas of maximized hazard. We can then use the decoder \dec\ to produce new samples $\widetilde{\x}$ from this transformed latent distribution. 
In this case, the disentanglement of latent features induced by the factorized Gaussian prior leads to samples, where survival-irrelevant characteristics are left relatively unchanged and survival-relevant characteristics are changed notably. 
In other words, the \grahl{} allows practitioners to understand the model's learned relationships directly in the original sample space by, e.g., plotting images based on the most survival-relevant changes in the latent space. This procedure can also be reversed, using the negative expected gradient for the parameter updates to generate samples with minimal risk.


\section{Experiments}

We evaluate the \grahl{} procedure on two datasets. First, on a simulated dataset based on MNIST, for which survival times are generated as described in \cite{Gensheimer2019}. The result is an MNIST dataset with the original images, where high hazard rates correspond to a high digit and vice versa. Second, we demonstrate the proposed approach on a dataset of 492 CT scans of patients with liver metastases, where the survival time after imaging is known.

Our \enc\ and \dec\ each consist of four residual blocks \parencite{He2016} with downsampling or upsampling operations respectively. \cox\ was chosen to be a fully-connected network with no hidden layer and no bias, which equals a linear predictor. We use Adam \parencite{Kingma2015} as optimizer with a learning rate of $1\text{e-}4$ for $\theta, \phi$ and $1\text{e-}5$ for $\psi$.

\paragraph{\grahl{} on MNIST}

\begin{figure}[b]
     \centering
    \includegraphics[width=\textwidth]{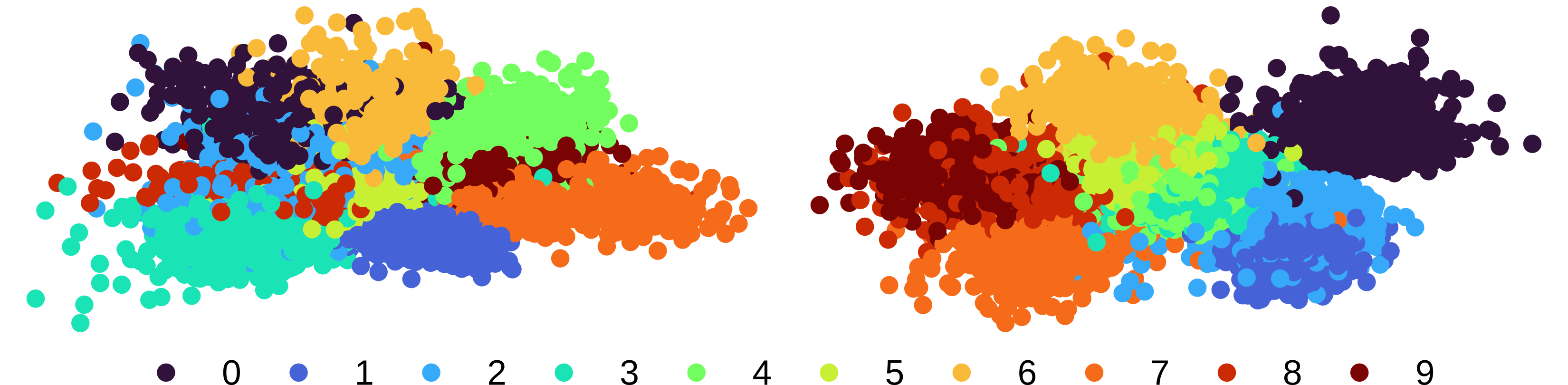}
    \caption{First two PCA components of MNIST embeddings from a VAE (\textbf{left}) versus a VAE with Cox objective (\textbf{right}). The color indicates the digit class. The VAE embedding is optimized to show structural similarities, whereas the Cox-regularized embedding groups classes according to their associated hazard.}
    \label{fig:mnist-embedding}
\end{figure}

Based on a latent space of 4 dimensions, Figure~\ref{fig:mnist-embedding} shows the differences in the embedding of a VAE with and without the regularizing Cox objective on the test dataset. As expected, the vanilla VAE (left plot) does not preserve the natural order of MNIST numbers, but structures the latent space with the intention of maximum visual reconstruction, hence the close neighborhood of digits like \texttt{0} and \texttt{9}. In contrast, when adding the survival information (right plot), the behavior changes. The main source of variation is now based on the actual hazard, i.e., the natural digit order. Digits such as \texttt{0} and \texttt{1} are now placed next to each other instead of being maximally separated as for the vanilla VAE due to their visual dissimilarity.

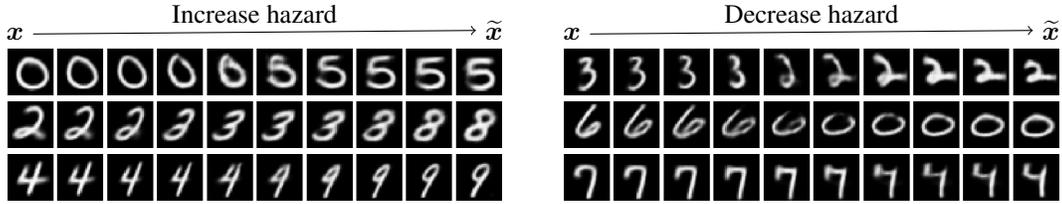
\begin{figure}[t]
     \centering
     \begin{subfigure}[b]{0.47\textwidth}
        \centering
    \begin{tikzpicture}[
         image/.style = {text width=\textwidth, 
                         inner sep=0pt, outer sep=0pt},
        node distance = 1mm and 1mm
                                ] 
        \node [image] (frame1)
            {\includegraphics[width=\textwidth]{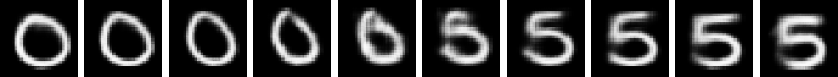}};
        \node [image,below=of frame1] (frame2) 
            {\includegraphics[width=\textwidth]{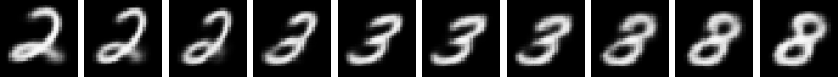}};
        \node[image,below=of frame2] (frame3)
            {\includegraphics[width=\textwidth]{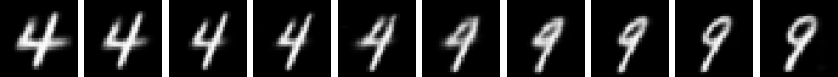}};
            
        \node [above=of frame1.west, yshift=2mm, xshift=1mm] (text1) {\x};
        \node [above=of frame1.east, yshift=2mm, xshift=-1mm] (text2) {$\widetilde{\x}$};
        
        \draw [->] (text1) -- node [above,midway] {Increase hazard} (text2);
        
    \end{tikzpicture}
        \label{fig:mnist-inc-walk}
     \end{subfigure}
     \hfill
    \begin{subfigure}[b]{0.47\textwidth}
        \centering
    \begin{tikzpicture}[
         image/.style = {text width=\textwidth, 
                         inner sep=0pt, outer sep=0pt},
        node distance = 1mm and 1mm
                                ] 
        \node [image] (frame1)
            {\includegraphics[width=\textwidth]{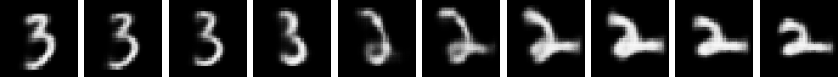}};
        \node [image,below=of frame1] (frame2) 
            {\includegraphics[width=\textwidth]{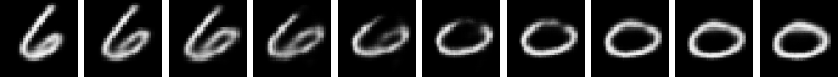}};
        \node[image,below=of frame2] (frame3)
            {\includegraphics[width=\textwidth]{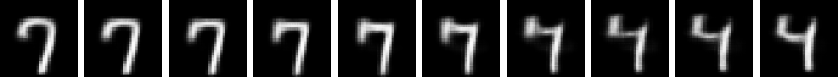}};
            
        \node [above=of frame1.west, yshift=2mm, xshift=1mm] (text1) {\x};
        \node [above=of frame1.east, yshift=2mm, xshift=-1mm] (text2) {$\widetilde{\x}$};
        
        \draw [->] (text1) -- node [above,midway] {Decrease hazard} (text2);
        
    \end{tikzpicture} \label{fig:mnist-inc-walk2}
     \end{subfigure}
     
     \caption{\grahl{} applied on three exemplary MNIST samples when increasing (\textbf{left}) or decreasing (\textbf{right}) the hazard rate. Applying the transformation results in a gradual change of the digit class while maintaining characteristics of the original sample \x.}
    \label{fig:mnist-walk}
\end{figure}

In a second step we investigate the effect of the \grahl{} on these survival-oriented embeddings. We apply the \grahl{} with 1500 iterations, 128 samples and visualize the progress of the transformation.
As depicted in Figure~\ref{fig:mnist-walk}, the gradients of \cox\ lead to a transition into different digit classes with higher (or lower) hazard. It is noteworthy that each digit shifts into a class that allows to keep some of the original core features (e.g., a \texttt{2} turns into a \texttt{3} first, then into an \texttt{8} instead of directly transforming into a \texttt{9} with the maximum associated hazard rate). 

\paragraph{\grahl{} on computed tomography imaging data}

CT imaging is an important step in the diagnostic workup of cancer patients and contains prognostic information. Here, we used axial, contrast-enhanced portal-venous CT images of 492 patients with hepatic metastases of colorectal cancer. Images were resampled to $1\text{mm}^3$ isotropic spatial resolution and liver as well as hepatic tumors were segmented using a pretrained nnU-Net \parencite{Isensee2018a}. For further computations, the volume was centered on the liver segmentation, cropped to $240^3$ voxels and downscaled to a resolution of $64^3$ voxels. On this volume,  the Cox-regularized VAE was trained with a latent space of 8. The input to the network consists  of the actual masked CT data and the segmentation of the tumor, represented as two channels.

A major challenge for model training is the small sample size in combination with a relatively high dimensional input space. This results in gaps in the otherwise smooth latent space.
\begin{figure}[h]
     \centering
     \begin{subfigure}[b]{0.85\textwidth}
        \centering
    \begin{tikzpicture}[
         image/.style = {text width=\textwidth, 
                         inner sep=0pt, outer sep=0pt},
        node distance = 1mm and 1mm
                                ] 
        \node [image] (frame1)
            {\includegraphics[width=\textwidth]{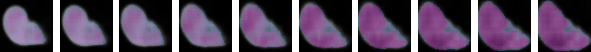}};
        \node [image,below=of frame1] (frame2) 
            {\includegraphics[width=\textwidth]{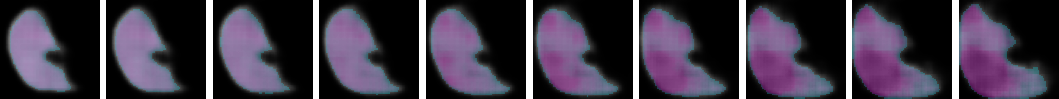}};
        \node[image,below=of frame2] (frame3)
            {\includegraphics[width=\textwidth]{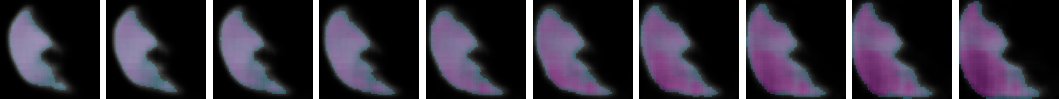}};
            
        \node [above=of frame1.west, yshift=4mm, xshift=1mm] (text1) {\x};
        \node [above=of frame1.east, yshift=4mm, xshift=-1mm] (text2) {$\widetilde{\x}$};
        
        \draw [->] (text1) -- node [above,midway] {Increase hazard} (text2);
        
    \end{tikzpicture}
        \caption{Increasing the hazard rate leads to a spread of (larger) tumor patches.}
        \label{fig:ctdata-inc-walk}
     \end{subfigure}
     \hfill
    \begin{subfigure}[b]{0.85\textwidth}
        \centering
    \begin{tikzpicture}[
         image/.style = {text width=\textwidth, 
                         inner sep=0pt, outer sep=0pt},
        node distance = 1mm and 1mm
                                ] 
        \node [image] (frame1)
            {\includegraphics[width=\textwidth]{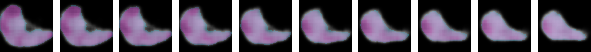}};
        \node [image,below=of frame1] (frame2) 
            {\includegraphics[width=\textwidth]{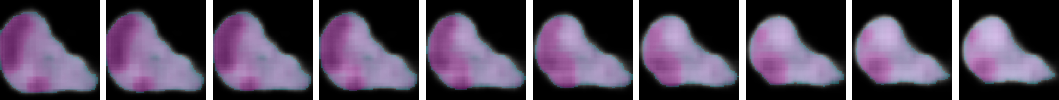}};
        \node[image,below=of frame2] (frame3)
            {\includegraphics[width=\textwidth]{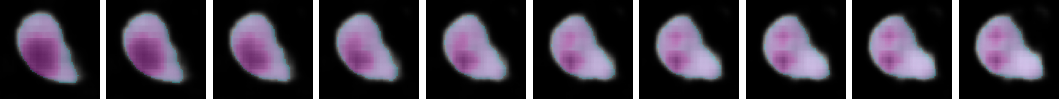}};
            
        \node [above=of frame1.west, yshift=4mm, xshift=1mm] (text1) {\x};
        \node [above=of frame1.east, yshift=4mm, xshift=-1mm] (text2) {$\widetilde{\x}$};
        
        \draw [->] (text1) -- node [above,midway] {Decrease hazard} (text2);
        
    \end{tikzpicture}
    \caption{Decreasing the hazard rate results in shrinking and vanishing tumor patches.}
    \label{fig:ctdata-inc-walk2}
     \end{subfigure}
     
     \caption{\grahl{} applied on computed tomography of the liver when increasing (\textbf{top}) or decreasing (\textbf{bottom}) the hazard rate with \grahl. The images show reconstructed liver CTs in grayscale with red-tinted tumor patches.}
    \label{fig:exp-ctdata-haz-walk-sample}
\end{figure}
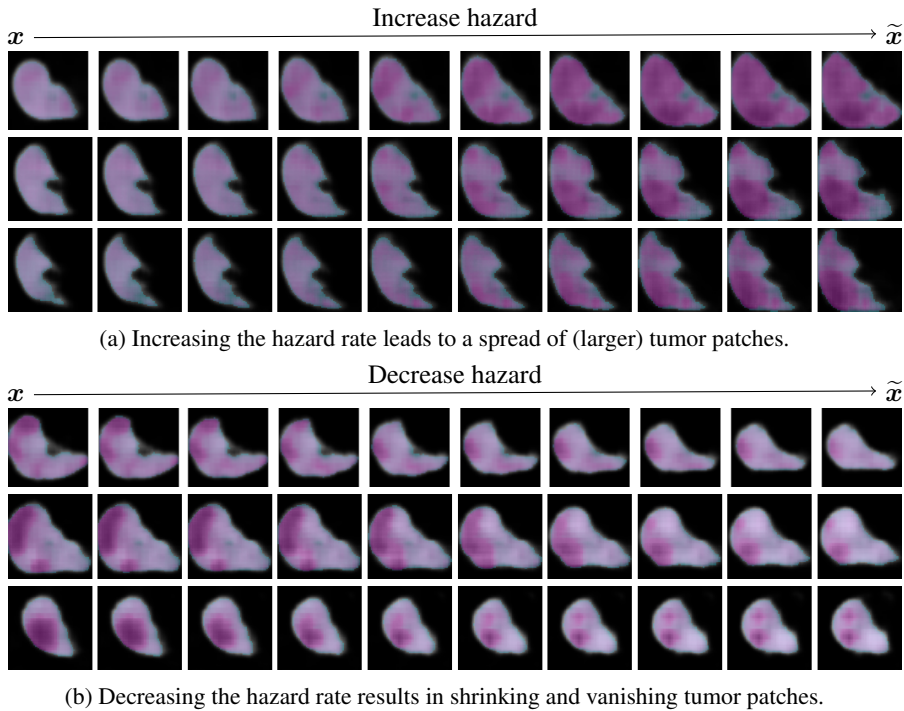
By increasing the $\beta$-parameter and therefore the influence of the $\mathbb{KL}$-term,
this can be mitigated at the cost of lower-quality reconstructions.
Figure~\ref{fig:exp-ctdata-haz-walk-sample} depicts multiple examples, showing different liver cancer cases where the hazard ratio is tweaked by \grahl{}.
Our method confirms the meaningfulness of the learned latent space of the autoencoder, increasing number and sizes of tumor patches when increasing hazard.
The shape of the liver changes only slightly, which indicates that overall survival may be unrelated to this characteristic.
In the given example, the sample in the first row of Figure~\ref{fig:ctdata-inc-walk} has a time-to-event of 1479 days and an estimated hazard rate of 0.601. This constitutes a hazard that is significantly lower than the baseline. After the transformation, the newly obtained $\widetilde{\x}$ is estimated with a hazard rate of 3.424.
On the contrary, reducing the hazard leads to exactly the opposite -- the tumor shrinks or vanishes. This can be seen, e.g., for the first sample in Figure~\ref{fig:ctdata-inc-walk2}, with an original time-to-event of 61 days and a hazard rate of 3.165 that transformed to 0.785.


\section{Conclusion and Outlook}


In this work we proposed a novel two-step procedure to learn and interpret the relationship between unstructured data sources such as images and survival times. While our generative model allows to emphasize hazard factors in unstructured data spaces, the proposed method \grahl{} yields a straightforward interpretation of the model using a gradient-based latent interpolation and reconstruction over the generative decoder.

One limiting factor in training generative models for SA are the typically small-scale datasets. 
This results in gaps in the latent space and/or insufficient representations.
Adjusting the $\mathbb{KL}$-term turned out to be beneficial in our experiments to focus on the quality of the representation, resulting in more reliable and stable \grahl{} interpolations.
While there are larger CT datasets available (e.g., \cite{Antonelli2021}), these are mostly for segmentation tasks and not associated with a survival outcome. An extension of our approach to a semi-supervised setting could thus allow the integration of more information to improve the encoding and generation. 
Alternatively, adversarial autoencoders as in \parencite{Larsen2015, Pidhorskyi2020} could help to improve the quality of generated images. Given additional tabular data, a possible direction for the VAE's downstream task could also be a semi-structured model \cite{Ruegamer2021}.

\section*{Acknowledgments}

This work has been partially supported by the German Federal Ministry of Education and Research (BMBF) under Grant No. 01IS18036A.
We thank the anonymous reviewers for their constructive comments, which helped us to improve the manuscript.

\printbibliography

\end{document}